\def\year{2018}\relax
\documentclass[letterpaper]{article} %DO NOT CHANGE THIS
\usepackage{aaai18}  %Required
\usepackage{times}  %Required
\usepackage{helvet}  %Required
\usepackage{courier}  %Required
\usepackage{url}  %Required
\usepackage{graphicx}  %Required

\usepackage{subcaption}
\graphicspath{{images/}}
\usepackage{amsmath}
\usepackage{amssymb}
\usepackage{multirow}
\usepackage{comment}

\begin{document}
% The file aaai.sty is the style file for AAAI Press 
% proceedings, working notes, and technical reports.
%
\title{Deep Reinforcement Learning for Unsupervised Video Summarization with Diversity-Representativeness Reward}
\author{Kaiyang Zhou,\textsuperscript{1,2} Yu Qiao\textsuperscript{1}\thanks{Corresponding author.}, Tao Xiang\textsuperscript{2} \\
\textsuperscript{1} Guangdong Key Lab of Computer Vision and Virtual Reality, \\ 
Shenzhen Institutes of Advanced Technology, Chinese Academy of Sciences, China \\
\textsuperscript{2} Queen Mary University of London, UK \\
k.zhou@qmul.ac.uk, yu.qiao@siat.ac.cn, t.xiang@qmul.ac.uk \\
}
\maketitle

%%%%%%%%% ABSTRACT
\begin{abstract}
Video summarization aims to facilitate large-scale video browsing by producing short, concise summaries that are diverse and representative of original videos. In this paper, we formulate video summarization as a sequential decision-making process and develop a deep summarization network (DSN) to summarize videos. DSN predicts for each video frame a probability, which indicates how likely a frame is selected, and then takes actions based on the probability distributions to select frames, forming video summaries. To train our DSN, we propose an end-to-end, reinforcement learning-based framework, where we design a novel reward function that \emph{jointly} accounts for diversity and representativeness of generated summaries and does not rely on labels or user interactions at all. During training, the reward function judges how diverse and representative the generated summaries are, while DSN strives for earning higher rewards by learning to produce more diverse and more representative summaries. Since labels are not required, our method can be \emph{fully unsupervised}. Extensive experiments on two benchmark datasets show that our unsupervised method not only outperforms other state-of-the-art unsupervised methods, but also is comparable to or even superior than most of published supervised approaches.
\end{abstract}

\section{Introduction}
Driven by the exponential growth in the amount of online videos in recent years, research in video summarization has gained increasing attention, leading to various methods proposed to facilitate large-scale video browsing \cite{gygli2014creating,gygli2015video,zhang2016summary,song2015tvsum,panda2017collaborative,mahasseniunsupervised,potapov2014category}.

Recently, recurrent neural network (RNN), especially with the long short-term memory (LSTM) cell \cite{hochreiter1997long}, has been exploited to model the sequential patterns in video frames, as well as to tackle the end-to-end training problem. Zhang et al. \cite{zhang2016video} proposed a deep architecture that combines a bidirectional LSTM network with a Determinantal Point Process (DPP) module that increases diversity in summaries, referring to as DPP-LSTM. They trained DPP-LSTM with supervised learning, using both video-level summaries and frame-level importance scores. At test time, DPP-LSTM predicts importance scores and outputs feature vectors simultaneously, which are together used to construct a DPP matrix. Due to the DPP modeling, DPP-LSTM needs to be trained in a two-stage manner.

Although DPP-LSTM \cite{zhang2016video} has shown state-of-the-art performances on several benchmarks, we argue that supervised learning cannot fully explore the potential of deep networks for video summarization because there does not exist a single ground truth summary for a video. This is grounded by the fact that humans have subjective opinions on which parts of a video should be selected as the summary. Therefore, devising more effective summarization methods that rely less on labels is still in demand.

Mahasseni et al. \cite{mahasseniunsupervised} developed an adversarial learning framework to train DPP-LSTM. During the learning process, DPP-LSTM selects keyframes and a discriminator network is used to judge whether a synthetic video constructed by the keyframes is real or not, in order to enforce DPP-LSTM to select more representative frames. Although their framework is unsupervised, the adversarial nature makes the training unstable, which may result in model collapse. In terms of increasing diversity, DPP-LSTM cannot benefit maximally from the DPP module without the help of labels. Since a RNN-based encoder-decoder network following DPP-LSTM for video reconstruction requires pretraining, their framework requires multiple training stages, which is not efficient in practice.

In this paper, we formulate video summarization as a sequential decision-making process and develop a deep summarization network (DSN) to summarize videos. DSN has an encoder-decoder architecture, where the encoder is a convolutional neural network (CNN) that performs feature extraction on video frames and the decoder is a bidirectional LSTM network that produces probabilities based on which actions are sampled to select frames. To train our DSN, we propose an end-to-end, reinforcement learning-based framework with a diversity-representativeness (DR) reward function that jointly accounts for diversity and representativeness of generated summaries, and does not rely on labels or user interactions at all.

The DR reward function is inspired by the general criteria of what properties a high-quality video summary should have. Specifically, the reward function consists of a diversity reward and a representativeness reward. The diversity reward measures how dissimilar the selected frames are to each other, while the representativeness reward computes distances between frames and their nearest selected frames, which is essentially the k-medoids problem. These two rewards complement to each other and work jointly to encourage DSN to produce diverse, representative summaries. The intuition behind this learning strategy is closely concerned with how humans summarize videos. To the best of our knowledge, this paper is the first to apply reinforcement learning to unsupervised video summarization.

The learning objective of DSN is to maximize the expected rewards over time. The rationale for using reinforcement learning (RL) to train DSN is two-fold. Firstly, we use RNN as part of our model and focus on the unsupervised setting. RNN needs to receive supervision signals at each temporal step but our rewards are computed over the whole video sequence, i.e., they can only be obtained after a sequence finishes. To provide supervision from a reward that is only available in the end of sequence, RL becomes a natural choice. Secondly,  we conjecture that DSN can benefit more from RL because RL essentially aims to optimize the action (frame-selection) mechanism of an agent by iteratively enforcing the agent to take better and better actions. However, optimizing action mechanism is not particularly highlighted in a normal supervised/unsupervised setting.

As the training process does not require labels, our method can be fully unsupervised. To fit the case where labels are available, we further extend our unsupervised method to the supervised version by adding a supervised objective that directly maximizes the log-probability of selecting annotated keyframes. By learning the high-level concepts encoded in labels, our DSN can recognize globally important frames and produce summaries that highly align with human-annotated summaries.

We conduct extensive experiments on two datasets, SumMe \cite{gygli2014creating} and TVSum \cite{song2015tvsum}, to quantitatively and qualitatively evaluate our method. The quantitative results show that our unsupervised method not only outperforms other state-of-the-art unsupervised alternatives, but also is comparable to or even superior than most of published supervised methods. More impressively, the qualitative results illustrate that DSN trained with our unsupervised learning algorithm can identify important frames that coincide with human selections.

The main contributions of this paper are summarized as follows: (1) We develop an end-to-end, reinforcement learning-based framework for training DSN, where we propose a label-free reward function that jointly accounts for diversity and representativeness of generated summaries. To the best of our knowledge, our work is the first to apply reinforcement learning to unsupervised video summarization. (2) We extend our unsupervised approach to the supervised version to leverage labels. (3) We conduct extensive experiments on two benchmark datasets to show that our unsupervised method not only outperforms other state-of-the-art unsupervised methods, but also is comparable to or even superior than most of published supervised approaches.

\section{Related Work}
\begin{figure*}[h]
\centering
\includegraphics[width=0.95\textwidth]{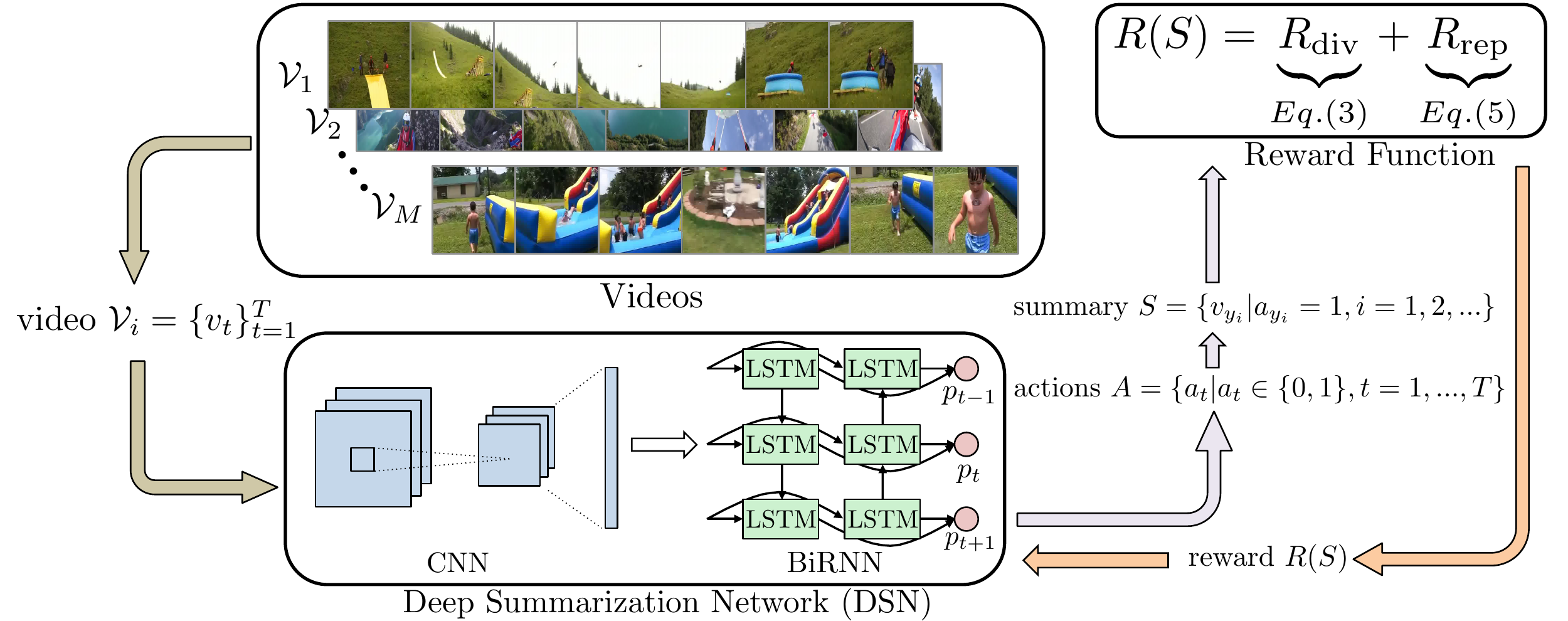}
\caption{Training deep summarization network (DSN) via reinforcement learning. DSN receives a video $\mathcal{V}_i$ and takes actions $A$ (i.e., a sequence of binary variables) on which parts of the video are selected as the summary $S$. The feedback reward $R(S)$ is computed based on the quality of the summary, i.e., diversity and representativeness.}
\label{fig:pipeline}
\end{figure*}

{\bf Video summarization.}
Research in video summarization has been significantly advanced in recent years, leading to approaches of various characteristics. Lee et al. \cite{lee2012discovering} identified important objects and people in summarizing videos. Gygli et al. \cite{gygli2014creating} learned a linear regressor to predict the degree of interestingness of video frames and selected keyframes with the highest interestingness scores. Gygli et al. \cite{gygli2015video} cast video summarization as a subset selection problem and optimized submodular functions with multiple objectives. Ejaz et al. \cite{ejaz2013efficient} applied an attention-modeling technique to extracting keyframes of visual saliency. Zhang et al. \cite{zhang2016summary} developed a nonparametric approach to transfer structures of known video summaries to new videos with similar topics. Auxiliary resources have also been exploited to aid the summarization process such as web images/videos \cite{song2015tvsum,khosla2013large,chu2015video} and category information \cite{potapov2014category}. Most of these non-deep summarization methods processed video frames independently, thus ignoring the inherent sequential patterns. Moreover, non-deep summarization methods usually do not support end-to-end training, which causes extra costs at test time. To address the aforementioned issues, we model video summarization via a deep RNN to capture long-term dependencies in video frames, and propose a reinforcement learning-based framework to train the network end to end.

{\bf Reinforcement learning (RL).}
RL has become an increasingly popular research area due to its effectiveness in various tasks. Mnih et al. \cite{mnih2013playing} successfully approximated Q function with a deep CNN, and enabled their agent to beat a human expert in several Atari games. Later on, many researchers have applied RL algorithms to vision-related applications such as image captioning \cite{xu2015show} and person re-identification \cite{xulan2017reid}. In the domain of video summarization, our work is not the first to use RL. Previously, Song et al. \cite{song2016category} has applied RL to training a summarization network for selecting category-specific keyframes. Their learning framework requires keyframe-labels and category information of training videos. However, our work significantly differs from the work of Song et al. and other RL-based work in the way that labels or user interactions are not required at all during the learning process, which is attributed to our novel reward function. Therefore, our summarization method can be fully unsupervised and is more practical to be deployed for large-scale video summarization.

\section{Proposed Approach}
We formulate video summarization as a sequential decision-making process. In particular, we develop a deep summarization network (DSN) to predict probabilities for video frames and make decisions on which frames to select based on the predicted probability distributions. We present an end-to-end, reinforcement learning-based framework for training our DSN, where we design a diversity-representativeness reward function, which directly assesses how diverse and representative the generated summaries are. Figure \ref{fig:pipeline} illustrates the overall learning process.

\subsection{Deep Summarization Network}
We adopt the encoder-decoder framework for our deep summarization network (DSN). The encoder is a convolutional neural network (CNN) that extracts visual features $\{ x_t \}_{t=1}^T$ from the input video frames $\{ v_t \}_{t=1}^T$ with the length $T$. The decoder is a bidirectional recurrent neural network (BiRNN) topped with a fully connected (FC) layer. The BiRNN takes as input the entire visual features $\{ x_t \}_{t=1}^T$ and produces corresponding hidden states $\{ h_t \}_{t=1}^T$. Each $h_t$ is the concatenation of the forward hidden state $h^f_t$ and the backward hidden state $h^b_t$, which encapsulate the future information and the past information with a strong emphasis on the parts surrounding the $t^{\text{th}}$ frame. The FC layer that ends with the sigmoid function predicts for each frame a probability $p_t$, from which a frame-selection action $a_t$ is sampled:
\begin{align}
p_t = \sigma (W h_t), \label{eq:fclayer} \\
a_t \sim \text{Bernoulli} (p_t),
\end{align}
where $\sigma$ represents the sigmoid function, $a_t \in \{ 0, 1 \}$ indicates whether the $t^{\text{th}}$ frame is selected or not. The bias in Eq.\,(\ref{eq:fclayer}) is omitted for brevity. A video summary is composed of the selected frames, $S = \{v_{y_i} | a_{y_i} = 1, i = 1, 2, ... \}$.

In practice, we use the GoogLeNet \cite{szegedy2015going} pretrained on ImageNet \cite{deng2009imagenet} as the CNN model. The visual feature vectors $\{ x_t \}_{t=1}^T$ are extracted from the penultimate layer of the GoogLeNet. For the RNN cell, we employ long short-term memory (LSTM) to enhance RNN's ability for capturing long-term dependencies in video frames. During training, we only update the decoder.

\subsection{Diversity-Representativeness Reward Function}
During training, DSN will receive a reward $R(S)$ that evaluates the quality of generated summaries, and the objective of DSN is to maximize the expected rewards over time by producing high-quality summaries. In general, a high-quality video summary is expected to be both diverse and representative of the original video so that temporal information across the entire video can be maximally preserved. To this end, we propose a novel reward that assesses the degree of diversity and representativeness of generated summaries. The proposed reward is composed of a diversity reward $R_\text{div}$ and a representativeness reward $R_\text{rep}$, which we detail as follows.

{\bf Diversity reward.} We evaluate the degree of diversity of a generated summary by measuring the dissimilarity among the selected frames in the feature space. Let the indices of the selected frames be $\mathcal{Y} = \{ y_i | a_{y_i} = 1, i = 1, ..., |\mathcal{Y}| \}$, we compute $R_\text{div}$ as the mean of the pairwise dissimilarities among the selected frames:
\begin{equation} \label{eq:diversityreward}
R_\text{div} = \frac{1}{|\mathcal{Y}| (|\mathcal{Y}| - 1)} \sum_{t \in \mathcal{Y}} \sum_{\substack{t^\prime \in \mathcal{Y} \\ t^\prime \neq t}} d (x_t, x_{t^\prime}),
\end{equation}
where $d(\cdot, \cdot)$ is the dissimilarity function calculated by
\begin{equation}
d (x_t, x_{t^\prime}) = 1 - \frac{x_t^T x_{t^\prime}}{||x_t||_2 ||x_{t^\prime}||_2}.
\end{equation}

Intuitively, the more diverse (or more dissimilar) the selected frames to each other, the higher the diversity reward that the agent can receive. However, Eq.\,(\ref{eq:diversityreward}) treats video frames as randomly permutable items which ignore the temporal structure inherent in sequential data. In fact, the similarity between two temporally distant frames should be ignored because they are essential to the storyline construction \cite{gong2014diverse}. To overcome this problem, we set $d (x_t, x_{t^\prime}) = 1$ if $|t -t^\prime| > \lambda$, where $\lambda$ controls the degree of temporal distance. We will validate this hypothesis in the Experiments section.

{\bf Representativeness reward.} This reward measures how well the generated summary can represent the original video. To this end, we formulate the degree of representativeness of a video summary as the k-medoids problem \cite{gygli2015video}. In particular, we want the agent to select a set of medoids such that the mean of squared errors between video frames and their nearest medoids is minimal. Therefore, we define $R_\text{rep}$ as
\begin{equation}
R_\text{rep} = \exp (- \frac{1}{T} \sum_{t=1}^T \min_{t^\prime \in \mathcal{Y}} || x_t - x_{t^\prime} ||_2).
\end{equation}

With this reward, the agent is encouraged to select frames that are close to the cluster centers in the feature space. An alternative formulation of $R_\text{rep}$ can be the inverse reconstruction errors achieved by the selected frames, but this formulation is too computationally expensive.

{\bf Diversity-representativeness reward.} $R_\text{div}$ and $R_\text{rep}$ complement to each other and work jointly to guide the learning of DSN:
\begin{equation} \label{eq:rewardfunction}
R(S) = R_\text{div} + R_\text{rep}.
\end{equation}

During training, $R_\text{div}$ and $R_\text{rep}$ are similar in the order of magnitude. In fact, it is non-trivial to keep $R_\text{div}$ and $R_\text{rep}$ at the same order of magnitude during training, thus none of them would dominate in gradient computation. We give zero reward to DSN when no frames are selected, i.e., the sampled actions are all zeros.

\subsection{Training with Policy Gradient}
The goal of our summarization agent is to learn a policy function $\pi_\theta$ with parameters $\theta$ by maximizing the expected rewards
\begin{equation}
J (\theta) = \mathbb{E}_{p_\theta (a_{1:T})} [R (S)],
\end{equation}
where $p_\theta (a_{1:T})$ denotes the probability distributions over possible action sequences, and $R(S)$ is computed by Eq.\,(\ref{eq:rewardfunction}). $\pi_\theta$ is defined by our DSN.

Following the REINFORCE algorithm proposed by Williams \cite{williams1992simple}, we can compute the derivative of the objective function $J (\theta)$ w.r.t. the parameters $\theta$ as
\begin{equation} \label{eq:basic_reinforce}
\triangledown_\theta J (\theta) = \mathbb{E}_{p_\theta (a_{1:T})} [R (S) \sum_{t=1}^T \triangledown_\theta \log \pi_\theta (a_t | h_t)],
\end{equation}
where $a_t$ is the action taken by DSN at time $t$ and $h_t$ is the hidden state from the BiRNN.

Since Eq.\,(\ref{eq:basic_reinforce}) involves the expectation over high-dimensional action sequences, which is hard to compute directly, we approximate the gradient by running the agent for $N$ episodes on the same video and then taking the average gradient
\begin{equation} \label{eq:approx_reinforce}
\triangledown_\theta J (\theta) \approx \frac{1}{N} \sum_{n=1}^N \sum_{t=1}^T R_n \triangledown_\theta \log \pi_\theta (a_t | h_t),
\end{equation}
where $R_n$ is the reward computed at the $n^{\text{th}}$ episode. Eq.\,(\ref{eq:approx_reinforce}) is also known as the episodic REINFORCE algorithm.

Although the gradient in Eq.\,(\ref{eq:approx_reinforce}) is a good estimate, it may contain high variance which will make the network hard to converge. A common countermeasure is to subtract the reward by a constant baseline $b$, so the gradient becomes
\begin{equation} \label{eq:reinforce_minus_baseline}
\triangledown_\theta J (\theta) \approx \frac{1}{N} \sum_{n=1}^N \sum_{t=1}^T (R_n - b) \triangledown_\theta \log \pi_\theta (a_t | h_t),
\end{equation}
where $b$ is simply computed as the moving average of rewards experienced so far for computational efficiency.

\subsection{Regularization}
Since selecting more frames will also increase the reward, we impose a regularization term on the probability distributions $p_{1:T}$ produced by DSN in order to constrain the percentage of frames selected for the summary. Inspired by \cite{mahasseniunsupervised}, we minimize the following term during training,
\begin{equation} \label{eq:lengthregterm}
L_\text{percentage} = || \frac{1}{T} \sum_{t=1}^T p_t - \epsilon ||^2,
\end{equation}
where $\epsilon$ determines the percentage of frames to be selected.

In addition, we also add the $\ell2$ regularization term on the weight parameters $\theta$ to avoid overfitting
\begin{equation} \label{eq:weightregterm}
L_\text{weight} = \sum_{i,j} \theta_{i,j}^2.
\end{equation}

\subsection{Optimization}
We optimize the policy function's parameters $\theta$ via stochastic gradient-based method. By combing the gradients computed from Eq.\,(\ref{eq:reinforce_minus_baseline}), Eq.\,(\ref{eq:lengthregterm}) and Eq.\,(\ref{eq:weightregterm}), we update $\theta$ as
\begin{equation} \label{eq:paramupdate}
\theta = \theta - \alpha \triangledown_\theta (- J + \beta_1 L_\text{percentage} + \beta_2 L_\text{weight}),
\end{equation}
where $\alpha$ is learning rate, and $\beta_1$ and $\beta_2$ are hyperparameters that balance the weighting.

In practice, we use Adam \cite{kingma2014adam} as the optimization algorithm. As a result of learning, the log-probability of actions taken by the network that have led to high rewards is increased, while that of actions that have resulted in low rewards is decreased.

\subsection{Extension to Supervised Learning}
Given the keyframe indices for a video, $\mathcal{Y}^* = \{ y^*_i | i = 1, ..., |\mathcal{Y}^*| \}$, we use Maximum Likelihood Estimation (MLE) to maximize the log-probability of selecting keyframes specified by $\mathcal{Y}^*$, $\log p(t; \theta)$ where $t \in \mathcal{Y}^*$. $p(t; \theta)$ is computed from Eq.\,(\ref{eq:fclayer}). The objective is formalized as
\begin{equation} \label{eq:supervisedmle}
L_\text{MLE} = \sum_{t \in \mathcal{Y}^*} \log p(t; \theta).
\end{equation}

\subsection{Summary Generation}
For a test video, we apply a trained DSN to predict the frame-selection probabilities as importance scores. We compute shot-level scores by averaging frame-level scores within the same shot. For temporal segmentation, we use KTS proposed by \cite{potapov2014category}. To generate a summary, we select shots by maximizing the total scores while ensuring that the summary length does not exceed a limit, which is usually $15\%$ of the video length. The maximization step is essentially the 0/1 Knapsack problem, which is known as NP-hard. We obtain a near-optimal solution via dynamic programming \cite{song2015tvsum}.

Besides evaluating generated summaries in the Experiments part, we also qualitatively analyze the raw predictions of DSN so as to exclude the effect of this summary generation step, by which we can better understand what DSN has learned.

\section{Experiments}
\subsection{Experimental Setup}
{\bf Datasets.} We evaluate our methods on SumMe \cite{gygli2014creating} and TVSum \cite{song2015tvsum}. SumMe consists of 25 user videos covering various topics such as holidays and sports. Each video in SumMe ranges from 1 to 6 minutes and is annotated by 15 to 18 persons, thus there are multiple ground truth summaries for each video. TVSum contains 50 videos, which include the topics of news, documentaries, etc. The duration of each video varies from 2 to 10 minutes. Similar to SumMe, each video in TVSum has 20 annotators that provide frame-level importance scores. Following \cite{song2015tvsum,zhang2016video}, we convert importance scores to shot-based summaries for evaluation. In addition to these two datasets, we exploit two other datasets, OVP\footnote{Open video project: https://open-video.org/.} that has 50 videos and YouTube \cite{de2011vsumm} that has 39 videos excluding cartoon videos, to evaluate our method in the settings where training data is augmented \cite{zhang2016video,mahasseniunsupervised}.

{\bf Evaluation metric.} For fair comparison with other approaches, we follow the commonly used protocol from \cite{zhang2016video} to compute F-score as the metric to assess the similarity between automatic summaries and ground truth summaries. We also follow \cite{zhang2016video} to deal with multiple ground truth summaries.

{\bf Evaluation settings.} We use three settings as suggested in \cite{zhang2016video} to evaluate our method. (1) Canonical: we use the standard 5-fold cross validation (5FCV), i.e., 80$\%$ of videos for training and the rest for testing. (2) Augmented: we still use the 5FCV but we augment the training data in each fold with OVP and YouTube. (3) Transfer: for a target dataset, e.g. SumMe or TVSum, we use the other three datasets as the training data to test the transfer ability of our model. 

{\bf Implementation details.} We downsample videos by 2 fps as did in \cite{zhang2016video}. We set the temporal distance $\lambda$ to 20, the $\epsilon$ in Eq.\,\ref{eq:lengthregterm} to 0.5, and the number of episodes $N$ to 5. The other hyperparameters $\alpha$, $\beta_1$ and $\beta_2$ in Eq.\,(\ref{eq:paramupdate}) are optimized via cross-validation. We set the dimension of hidden state in the RNN cell to 256 throughout this paper. Training is stopped when it reaches a maximum number of epochs (60 in our case). Early stopping is executed when reward creases to increase for a period of time (10 epochs in our experiments). We implement our method based on Theano \cite{al2016theano}\footnote{Codes are available on https://github.com/KaiyangZhou/vsumm-reinforce}.

{\bf Comparison.} To compare with other approaches, we implement Uniform sampling, K-medoids and Dictionary selection \cite{elhamifar2012see} by ourselves. We retrieve results of other approaches including Video-MMR \cite{li2010multi}, Vsumm \cite{de2011vsumm}, Web image \cite{khosla2013large}, Online sparse coding \cite{zhao2014quasi}, Co-archetypal \cite{song2015tvsum}, Interestingness \cite{gygli2014creating}, Submodularity \cite{gygli2015video}, Summary transfer \cite{zhang2016summary}, Bi-LSTM and DPP-LSTM \cite{zhang2016video}, $\text{GAN}_\text{dpp}$ and $\text{GAN}_\text{sup}$ \cite{mahasseniunsupervised} from published papers. Due to space limit, we do not include these citations in tables.

\subsection{Quantitative Evaluation}
We first compare our method with several baselines that differ in learning objectives. Then, we compare our methods with current state-of-the-art unsupervised/supervised approaches in the three evaluation settings.

% Figures showing the qualitative results
\begin{figure*}[t]
\centering
  \begin{subfigure}[b]{0.99\textwidth}
  \includegraphics[width=\textwidth]{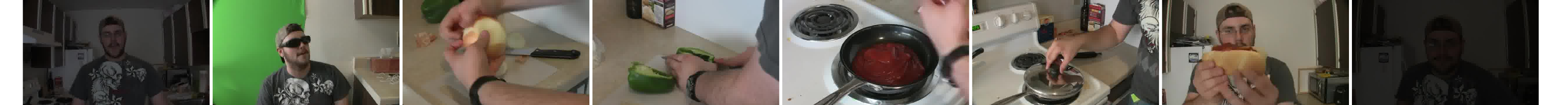}
  \caption{Example frames from video 18 in TVSum (indexed as in \cite{song2015tvsum}).}
  \label{fig:video18}
  \end{subfigure}
  ~
  \begin{subfigure}[b]{0.48\textwidth}
  \includegraphics[width=\textwidth]{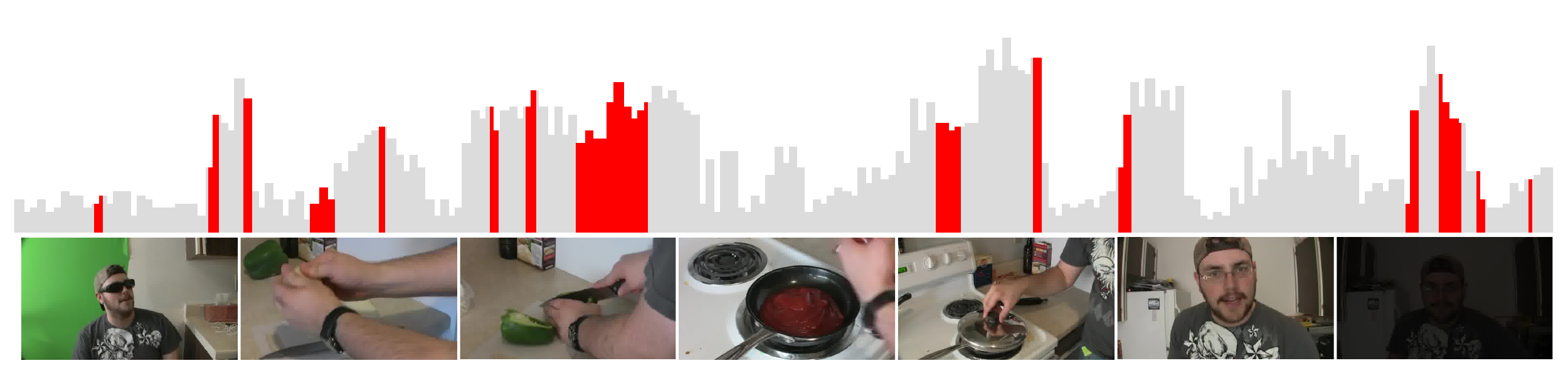}
  \caption{$\text{DR-DSN}_\text{sup}$}
  \label{fig:summDRRNNsup}
  \end{subfigure}
  ~
  \begin{subfigure}[b]{0.48\textwidth}
  \includegraphics[width=\textwidth]{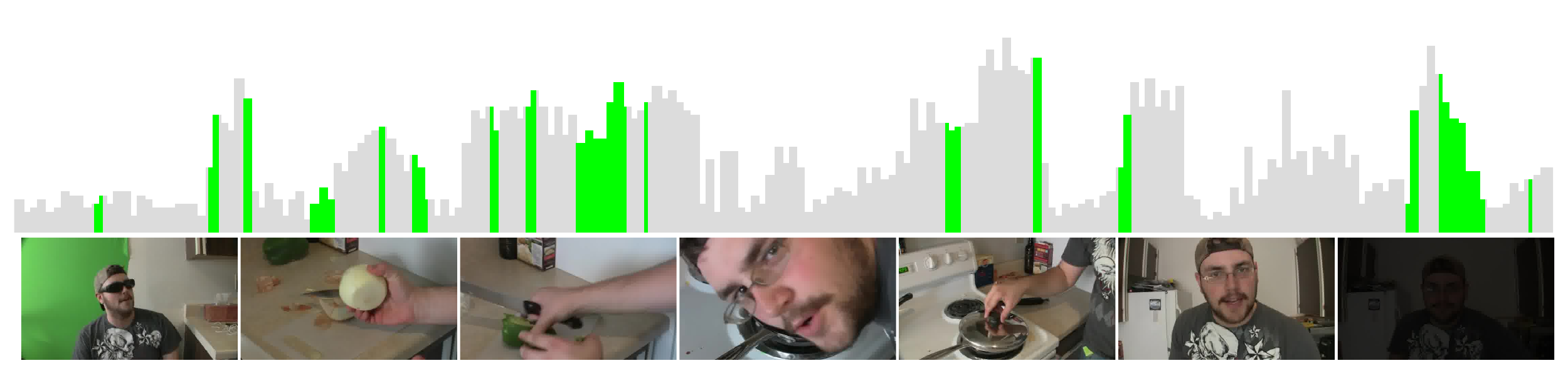}
  \caption{DR-DSN}
  \label{fig:summDRRNN}
  \end{subfigure}
  ~
  \begin{subfigure}[b]{0.48\textwidth}
  \includegraphics[width=\textwidth]{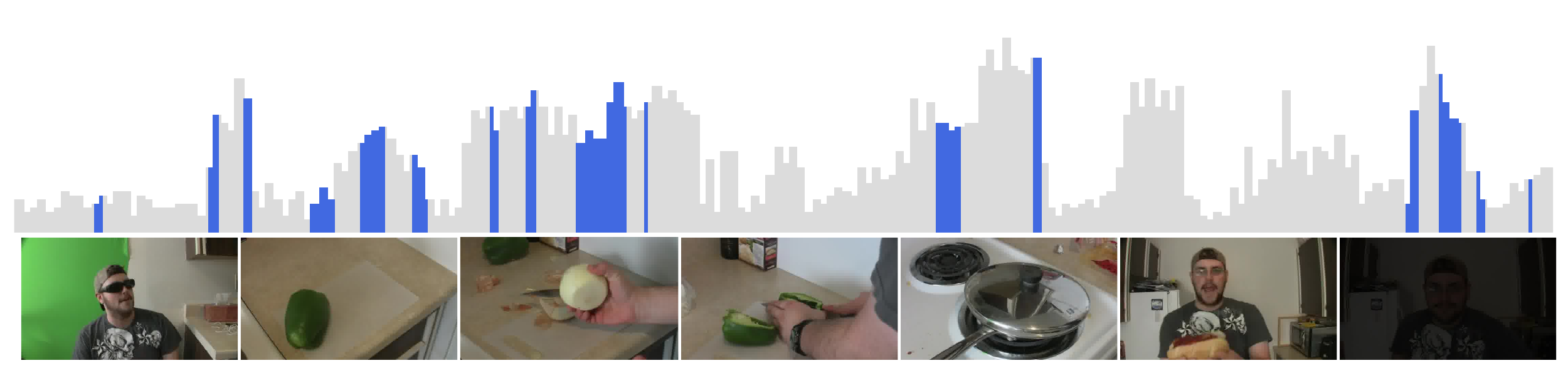}
  \caption{R-DSN}
  \label{fig:summRRNN}
  \end{subfigure}
  ~
  \begin{subfigure}[b]{0.48\textwidth}
  \includegraphics[width=\textwidth]{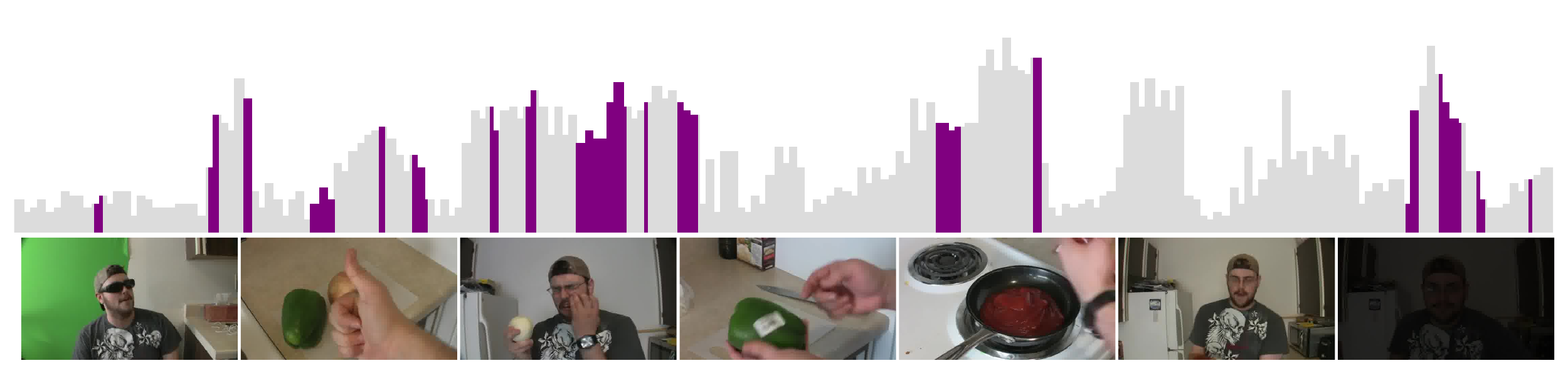}
  \caption{D-DSN}
  \label{fig:summDRNN}
  \end{subfigure}
\caption{Video summaries generated by different variants of our approach for video 18 in TVSum. The light-gray bars in (b) to (e) correspond to ground truth importance scores, while the colored areas correspond to the selected parts by different models.}
\label{fig:qualitativeresults}
\end{figure*}

{\bf Comparison with baselines.} We set the baseline models as the ones trained with $R_\text{div}$ only and $R_\text{rep}$ only, which are denoted by D-DSN and R-DSN, respectively. We represent the model trained with the two rewards jointly as DR-DSN. The model that is extended to the supervised version is denoted by $\text{DR-DSN}_\text{sup}$. We also validate the effectiveness of the proposed technique (we call this $\lambda$-technique from now on) that ignores the distant similarity when computing $R_\text{div}$. We represent the D-DSN trained without the $\lambda$-technique as $\text{D-DSN}_{\text{w/o $\lambda$}}$. To verify that DSN can benefit more from reinforcement learning than from supervised learning, we add another baseline as the DSN trained with the cross entropy loss using keyframe annotations, where a confidence penalty \cite{pereyra2017regularizing} is imposed on the output distributions as a regularization term. This model is denoted by $\text{DSN}_{\text{sup}}$.

\begin{table}[h]
\centering
\caption{Results ($\%$) of different variants of our method on SumMe and TVSum.}
\label{tb:comparebaselines}
\begin{tabular}{l | c | c}
\hline
\multicolumn{1}{c|}{Method} & SumMe & TVSum \\
\hline
$\text{DSN}_\text{sup}$ & 38.2 & 54.5 \\
$\text{D-DSN}_{\text{w/o $\lambda$}}$ & 39.3 & 55.7 \\
D-DSN & 40.5 & 56.2 \\
R-DSN & 40.7 & 56.9 \\
\hline \hline
DR-DSN & 41.4 & 57.6 \\
$\text{DR-DSN}_\text{sup}$ & {\bf 42.1} & {\bf 58.1} \\
\hline
\end{tabular}
\end{table}

Table \ref{tb:comparebaselines} reports the results of different variants of our method on SumMe and TVSum. We can see that DR-DSN clearly outperforms D-DSN and R-DSN on both datasets, which demonstrates that by using $R_\text{div}$ and $R_\text{rep}$ collaboratively, we can better teach DSN to produce high-quality summaries that are diverse and representative. Comparing the unsupervised model with the supervised one, we see that DR-DSN significantly outperforms $\text{DSN}_{\text{sup}}$ on the two datasets (41.4 vs. 38.2 on SumMe and 57.6 vs. 54.5 on TVSum), which justifies our assumption that DSN can benefit more from reinforcement learning than from supervised learning.

By adding the supervision signals of $L_\text{MLE}$ (Eq.\,(\ref{eq:supervisedmle})) to DR-DSN, the summarization performances are further improved (1.7$\%$ improvements on SumMe and 0.9$\%$ improvements on TVSum). This is because labels encode the high-level understanding of the video content, which is exploited by $\text{DR-DSN}_\text{sup}$ to learn more useful patterns. 

The performances of R-DSN are slightly better than those of D-DSN on the two datasets, which is because diverse summaries usually contain redundant information that are irrelevant to the video subject. We observe that the performances of D-DSN are better than those of $\text{D-DSN}_{\text{w/o $\lambda$}}$ that does not consider temporally distant frames. When using the $\lambda$-technique in training, around 50$\%$ $\sim$ 70$\%$ of the distance matrix was set to 1 (varying across different videos) at the early stage. As the training epochs increased, the percentage went up too, eventually staying around 80$\%$ $\sim$ 90$\%$. This makes sense because selecting temporally distant frames can lead to higher rewards and DSN is encouraged to do so with the diversity reward function.

{\bf Comparison with unsupervised approaches.} Table \ref{tb:resOfUnsup} shows the results of DR-DSN against other unsupervised approaches on SumMe and TVSum. It can be seen that DR-DSN outperforms the other unsupervised approaches on both datasets by large margins. On SumMe, DR-DSN is 5.9$\%$ better than the current state-of-the-art, $\text{GAN}_\text{dpp}$. On TVSum, DR-DSN substantially beats $\text{GAN}_\text{dpp}$ by 11.4$\%$. Although our reward functions are analogous to the objectives of $\text{GAN}_\text{dpp}$ in concepts, ours directly model diversity and representativeness of selected frames in the feature space, which is more useful to guide DSN to find good solutions. In addition, the training performances of DR-DSN are 40.2$\%$ on SumMe and 57.2$\%$ on TVSum, which suggest that the model did not overfit to the training data (note that we do not explicitly optimize the F-score metric in the training objective function).

\begin{table}[h]
\centering
\caption{Results ($\%$) of unsupervised approaches on SumMe and TVSum. Our DR-DSN performs the best, especially in TVSum where it exhibits a huge advantage over others.}
\label{tb:resOfUnsup}
\begin{tabular}{l | c | c}
\hline
\multicolumn{1}{c|}{Method} & SumMe & TVSum \\
\hline
Video-MMR & 26.6 & - \\
Uniform sampling & 29.3 & 15.5 \\
K-medoids & 33.4 & 28.8 \\
Vsumm & 33.7 & - \\
Web image & - & 36.0 \\
Dictionary selection & 37.8 & 42.0 \\
Online sparse coding & - & 46.0 \\
Co-archetypal & - & 50.0 \\
$\text{GAN}_\text{dpp}$ & 39.1 & 51.7 \\
\hline \hline
DR-DSN & {\bf 41.4} & {\bf 57.6} \\
\hline
\end{tabular}
\end{table}

\begin{figure*}[h]
\centering
  \begin{subfigure}[b]{0.49\textwidth}
  \includegraphics[width=\textwidth]{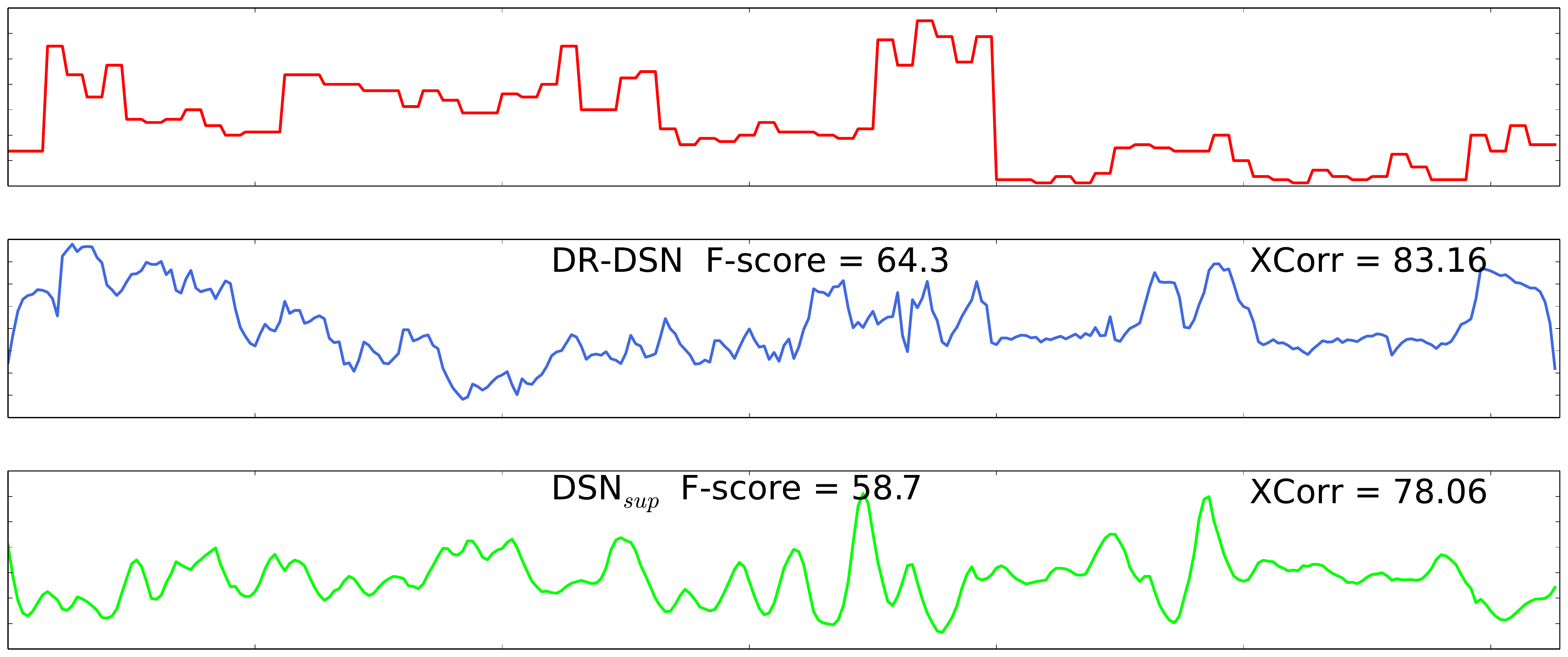}
  \caption{Video 11 in TVSum} \label{fig:tvsum_video_11}
  \end{subfigure}
  ~
  \begin{subfigure}[b]{0.49\textwidth}
  \includegraphics[width=\textwidth]{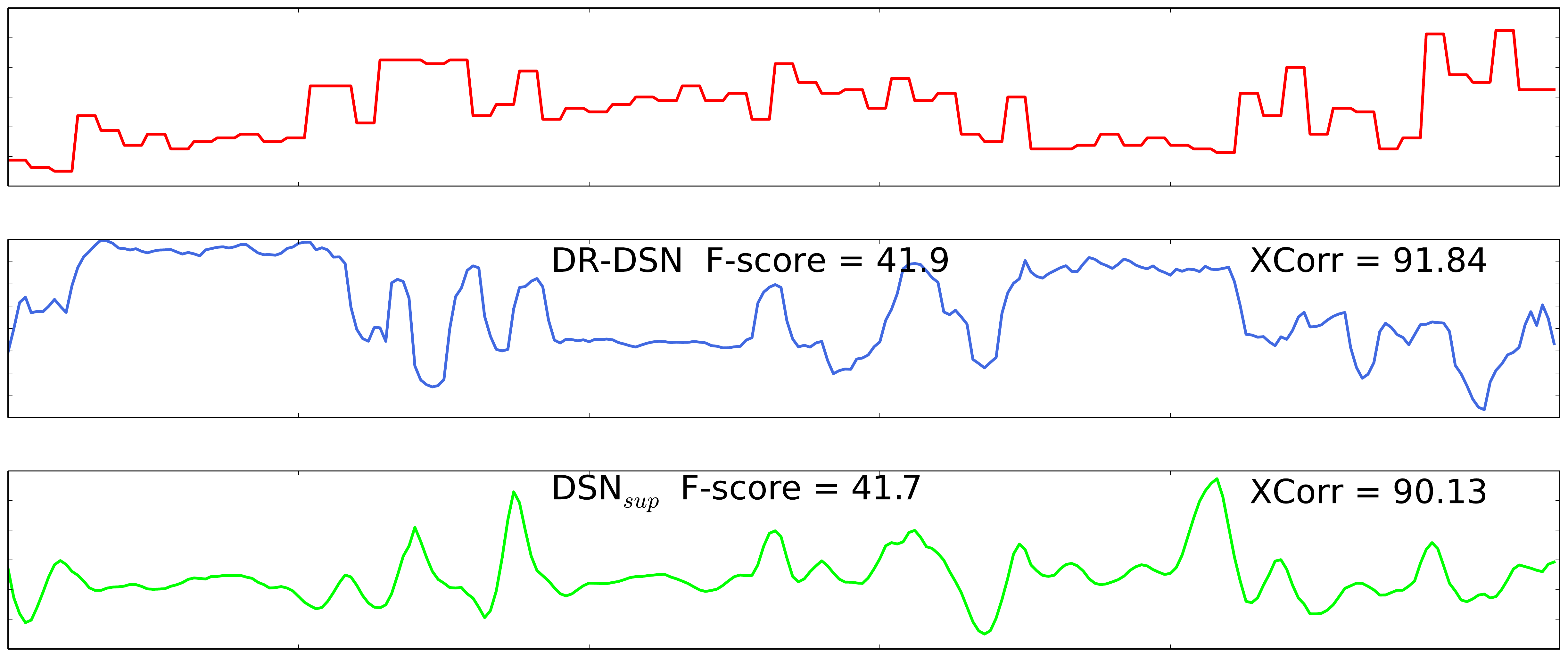}
  \caption{Video 10 in TVSum} \label{fig:tvsum_video_10}
  \end{subfigure}
\caption{Ground truth (top) and importance scores predicted by DR-DSN (middle) and $\text{DSN}_\text{sup}$ (bottom). Besides the F-score for each prediction, we also compute cross-correlation (XCorr) for each pair of prediction and ground truth to give a quantitative measure of similarity over two series of 1D arrays. The higher the XCorr, the more similar two arrays are to each other.}
\label{fig:IS_tvsum_videos}
\end{figure*}

{\bf Comparison with supervised approaches.} Table \ref{tb:resOfSup} reports the results of our supervised model, $\text{DR-DSN}_\text{sup}$, and other supervised approaches. In terms of LSTM-based methods, our $\text{DR-DSN}_\text{sup}$ beats the others, i.e., Bi-LSTM, DPP-LSTM and $\text{GAN}_\text{sup}$, by 1.0$\% \sim$ 12.0$\%$ on SumMe and 3.2$\% \sim$ 7.2$\%$ on TVSum, respectively. It is also interesting to see that the summarization performance of our unsupervised method, DR-DSN, is even superior than the state-of-the-art supervised approach on TVSum (57.6 vs. 56.3), and is better than most of the supervised approaches on SumMe. These results strongly prove the efficacy of our learning framework.

\begin{table}[h]
\centering
\caption{Results ($\%$) of supervised approaches on SumMe and TVSum. Our $\text{DR-DSN}_\text{sup}$ performs the best.}
\label{tb:resOfSup}
\begin{tabular}{l | c | c}
\hline
\multicolumn{1}{c|}{Method} & SumMe & TVSum \\
\hline
Interestingness & 39.4 & - \\
Submodularity & 39.7 & - \\
Summary transfer & 40.9 & - \\
Bi-LSTM & 37.6 & 54.2 \\
DPP-LSTM & 38.6 & 54.7 \\
$\text{GAN}_\text{sup}$ & 41.7 & 56.3 \\
\hline \hline
$\text{DR-DSN}_\text{sup}$ & {\bf 42.1} & {\bf 58.1} \\
\hline
\end{tabular}
\end{table}

{\bf Comparison in the Augmented (A) and Transfer (T) settings.} Table \ref{tb:resOfAugTran} compares our methods with current state-of-the-art LSTM-based methods in the A and T settings. The results in the Canonical setting are also provided to exhibit the improvements obtained by increased training data. In the A setting, $\text{DR-DSN}_\text{sup}$ performs marginally better than $\text{GAN}_\text{sup}$ on SumMe (43.9 vs. 43.6), whereas it is defeated by $\text{GAN}_\text{sup}$ on TVSum (59.8 vs. 61.2). This may be because the LSTM model in $\text{GAN}_\text{sup}$ has more hidden units (1024 vs. our 256). In the T setting, $\text{DR-DSN}_\text{sup}$ performs the best on both datasets, suggesting that our model is able to transfer knowledge between datasets. Furthermore, it is interesting to see that our unsupervised model, DR-DSN, is superior or comparable with other methods in both settings. Overall, we firmly believe that by using a larger model and/or designing a better network architecture, we can obtain better summarization performances with our learning framework.

\begin{table}[h]
\centering
\caption{Results ($\%$) of the LSTM-based approaches on SumMe and TVSum in the Canonical (C), Augmented (A) and Transfer (T) settings, respectively.}
\label{tb:resOfAugTran}
\begin{tabular}{l | c | c | c | c | c | c}
\hline
\multicolumn{1}{c|}{\multirow{2}{*}{Method}} & \multicolumn{3}{c|}{SumMe} & \multicolumn{3}{c}{TVSum} \\ \cline{2-7} 
\multicolumn{1}{c|}{} & C & A & T & C & A & T \\ 
\hline
Bi-LSTM & 37.6 & 41.6 & 40.7 & 54.2 & 57.9 & 56.9 \\
DPP-LSTM & 38.6 & 42.9 & 41.8 & 54.7 & 59.6 & 58.7 \\
$\text{GAN}_\text{dpp}$ & 39.1 & 43.4 & - & 51.7 & 59.5 & - \\
$\text{GAN}_\text{sup}$ & 41.7 & 43.6 & - & 56.3 & {\bf 61.2} & - \\
\hline \hline
DR-DSN & 41.4 & 42.8 & 42.4 & 57.6 & 58.4 & 57.8 \\
$\text{DR-DSN}_\text{sup}$ & {\bf 42.1} & {\bf 43.9} & {\bf 42.6} & {\bf 58.1} & 59.8 & {\bf 58.9} \\ 
\hline
\end{tabular}
\end{table}

We also experiment with different gated RNN units, i.e., LSTM vs. GRU \cite{cho2014properties}, and find that LSTM-based models consistently beat GRU-based models (see Table \ref{tb:lstmgru}). This may be interpreted as that the memory mechanism in LSTM has a higher degree of complexity, thus allowing more complex patterns to be learned.

\begin{table}[h]
\centering
\caption{Results ($\%$) of using different gated recurrent units.}
\label{tb:lstmgru}
\begin{tabular}{l | c | c | c | c}
\hline
\multicolumn{1}{c|}{\multirow{2}{*}{Method}} & \multicolumn{2}{c|}{SumMe} & \multicolumn{2}{c}{TVSum} \\ \cline{2-5} 
\multicolumn{1}{c|}{} & LSTM & GRU & LSTM & GRU \\ \hline
DR-DSN & 41.4 & 41.2 & 57.6 & 56.7 \\ \hline
$\text{DR-DSN}_\text{sup}$ & 42.1 & 41.5 & 58.1 & 57.8 \\ \hline
\end{tabular}
\end{table}

\subsection{Qualitative Evaluation}
{\bf Video summaries.} We provide qualitative results for an exemplar video that talks about a man making a spicy sausage sandwich in Figure \ref{fig:qualitativeresults}. In general, all four methods produce high-quality summaries that span the temporal structure, with only small variations observed in some frames. The peak regions of ground truth are almost captured. Nevertheless, the summary produced by the supervised model, $\text{DR-DSN}_\text{sup}$, is much closer to the complete storyline conveyed by the original video i.e., from food preparation to cooking. This is because $\text{DR-DSN}_\text{sup}$ benefits from labels that allow high-level concepts to be better captured.

{\bf Predicted importance scores.} We visualize the raw predictions by DR-DSN and $\text{DSN}_\text{sup}$ in Figure \ref{fig:IS_tvsum_videos}. By comparing predictions with ground truth, we can better understand in more depth how well DSN has learned. It is worth highlighting that the curves of importance scores predicted by the unsupervised model resemble those predicted by the supervised model in several parts. More importantly, these parts coincide with the ones also considered as important by humans. This strongly demonstrates that reinforcement learning with our diversity-representativeness reward function can well imitate the human-learning process and effectively teach DSN to recognize important frames.

\section{Conclusion}
In this paper, we proposed a label-free reinforcement learning algorithm to tackle unsupervised video summarization. Extensive experiments on two benchmark datasets showed that using reinforcement learning with our unsupervised reward function outperformed other state-of-the-art unsupervised alternatives, and produced results comparable to or even superior than most supervised methods.

\section{Acknowledgments}
We thank Ke Zhang and Wei-Lun Chao for discussions of details of their paper \cite{zhang2016video}. This work was supported in part by National Key Research and Development Program of China (2016YFC1400704) and 
National Natural Science Foundation of China (U1613211, 61633021).

\bibliography{vsummBiB}

\begin{thebibliography}{}

\bibitem[\protect\citeauthoryear{Al-Rfou \bgroup et al\mbox.\egroup
  }{2016}]{al2016theano}
Al-Rfou, R.; Alain, G.; Almahairi, A.; Angermueller, C.; Bahdanau, D.; Ballas,
  N.; Bastien, F.; Bayer, J.; Belikov, A.; Belopolsky, A.; et~al.
\newblock 2016.
\newblock Theano: A python framework for fast computation of mathematical
  expressions.
\newblock {\em arXiv preprint arXiv:1605.02688}.

\bibitem[\protect\citeauthoryear{Cho \bgroup et al\mbox.\egroup
  }{2014}]{cho2014properties}
Cho, K.; Van~Merri{\"e}nboer, B.; Bahdanau, D.; and Bengio, Y.
\newblock 2014.
\newblock On the properties of neural machine translation: Encoder-decoder
  approaches.
\newblock {\em arXiv preprint arXiv:1409.1259}.

\bibitem[\protect\citeauthoryear{Chu, Song, and Jaimes}{2015}]{chu2015video}
Chu, W.-S.; Song, Y.; and Jaimes, A.
\newblock 2015.
\newblock Video co-summarization: Video summarization by visual co-occurrence.
\newblock In {\em CVPR},  3584--3592.

\bibitem[\protect\citeauthoryear{De~Avila \bgroup et al\mbox.\egroup
  }{2011}]{de2011vsumm}
De~Avila, S. E.~F.; Lopes, A. P.~B.; da~Luz, A.; and de~Albuquerque~Ara{\'u}jo,
  A.
\newblock 2011.
\newblock Vsumm: A mechanism designed to produce static video summaries and a
  novel evaluation method.
\newblock {\em Pattern Recognition Letters} 32(1):56--68.

\bibitem[\protect\citeauthoryear{Deng \bgroup et al\mbox.\egroup
  }{2009}]{deng2009imagenet}
Deng, J.; Dong, W.; Socher, R.; Li, L.-J.; Li, K.; and Fei-Fei, L.
\newblock 2009.
\newblock Imagenet: A large-scale hierarchical image database.
\newblock In {\em CVPR},  248--255.
\newblock IEEE.

\bibitem[\protect\citeauthoryear{Ejaz, Mehmood, and
  Baik}{2013}]{ejaz2013efficient}
Ejaz, N.; Mehmood, I.; and Baik, S.~W.
\newblock 2013.
\newblock Efficient visual attention based framework for extracting key frames
  from videos.
\newblock {\em Signal Processing: Image Communication} 28(1):34--44.

\bibitem[\protect\citeauthoryear{Elhamifar, Sapiro, and
  Vidal}{2012}]{elhamifar2012see}
Elhamifar, E.; Sapiro, G.; and Vidal, R.
\newblock 2012.
\newblock See all by looking at a few: Sparse modeling for finding
  representative objects.
\newblock In {\em CVPR},  1600--1607.
\newblock IEEE.

\bibitem[\protect\citeauthoryear{Gong \bgroup et al\mbox.\egroup
  }{2014}]{gong2014diverse}
Gong, B.; Chao, W.-L.; Grauman, K.; and Sha, F.
\newblock 2014.
\newblock Diverse sequential subset selection for supervised video
  summarization.
\newblock In {\em NIPS},  2069--2077.

\bibitem[\protect\citeauthoryear{Gygli \bgroup et al\mbox.\egroup
  }{2014}]{gygli2014creating}
Gygli, M.; Grabner, H.; Riemenschneider, H.; and Van~Gool, L.
\newblock 2014.
\newblock Creating summaries from user videos.
\newblock In {\em ECCV},  505--520.
\newblock Springer.

\bibitem[\protect\citeauthoryear{Gygli, Grabner, and
  Van~Gool}{2015}]{gygli2015video}
Gygli, M.; Grabner, H.; and Van~Gool, L.
\newblock 2015.
\newblock Video summarization by learning submodular mixtures of objectives.
\newblock In {\em CVPR},  3090--3098.

\bibitem[\protect\citeauthoryear{Hochreiter and
  Schmidhuber}{1997}]{hochreiter1997long}
Hochreiter, S., and Schmidhuber, J.
\newblock 1997.
\newblock Long short-term memory.
\newblock {\em Neural computation} 9(8):1735--1780.

\bibitem[\protect\citeauthoryear{Khosla \bgroup et al\mbox.\egroup
  }{2013}]{khosla2013large}
Khosla, A.; Hamid, R.; Lin, C.-J.; and Sundaresan, N.
\newblock 2013.
\newblock Large-scale video summarization using web-image priors.
\newblock In {\em CVPR},  2698--2705.

\bibitem[\protect\citeauthoryear{Kingma and Ba}{2014}]{kingma2014adam}
Kingma, D., and Ba, J.
\newblock 2014.
\newblock Adam: A method for stochastic optimization.
\newblock In {\em ICLR}.

\bibitem[\protect\citeauthoryear{Lan \bgroup et al\mbox.\egroup
  }{2017}]{xulan2017reid}
Lan, X.; Wang, H.; Gong, S.; and Zhu, X.
\newblock 2017.
\newblock Deep reinforcement learning attention selection for person
  re-identification.
\newblock In {\em BMVC}.

\bibitem[\protect\citeauthoryear{Lee, Ghosh, and
  Grauman}{2012}]{lee2012discovering}
Lee, Y.~J.; Ghosh, J.; and Grauman, K.
\newblock 2012.
\newblock Discovering important people and objects for egocentric video
  summarization.
\newblock In {\em CVPR},  1346--1353.
\newblock IEEE.

\bibitem[\protect\citeauthoryear{Li and Merialdo}{2010}]{li2010multi}
Li, Y., and Merialdo, B.
\newblock 2010.
\newblock Multi-video summarization based on video-mmr.
\newblock In {\em WIAMIS},  1--4.
\newblock IEEE.

\bibitem[\protect\citeauthoryear{Mahasseni, Lam, and
  Todorovic}{2017}]{mahasseniunsupervised}
Mahasseni, B.; Lam, M.; and Todorovic, S.
\newblock 2017.
\newblock Unsupervised video summarization with adversarial lstm networks.
\newblock In {\em CVPR}.

\bibitem[\protect\citeauthoryear{Mnih \bgroup et al\mbox.\egroup
  }{2013}]{mnih2013playing}
Mnih, V.; Kavukcuoglu, K.; Silver, D.; Graves, A.; Antonoglou, I.; Wierstra,
  D.; and Riedmiller, M.
\newblock 2013.
\newblock Playing atari with deep reinforcement learning.
\newblock {\em arXiv preprint arXiv:1312.5602}.

\bibitem[\protect\citeauthoryear{Panda and
  Roy-Chowdhury}{2017}]{panda2017collaborative}
Panda, R., and Roy-Chowdhury, A.~K.
\newblock 2017.
\newblock Collaborative summarization of topic-related videos.
\newblock In {\em CVPR}.

\bibitem[\protect\citeauthoryear{Pereyra \bgroup et al\mbox.\egroup
  }{2017}]{pereyra2017regularizing}
Pereyra, G.; Tucker, G.; Chorowski, J.; Kaiser, {\L}.; and Hinton, G.
\newblock 2017.
\newblock Regularizing neural networks by penalizing confident output
  distributions.
\newblock {\em arXiv preprint arXiv:1701.06548}.

\bibitem[\protect\citeauthoryear{Potapov \bgroup et al\mbox.\egroup
  }{2014}]{potapov2014category}
Potapov, D.; Douze, M.; Harchaoui, Z.; and Schmid, C.
\newblock 2014.
\newblock Category-specific video summarization.
\newblock In {\em ECCV},  540--555.
\newblock Springer.

\bibitem[\protect\citeauthoryear{Song \bgroup et al\mbox.\egroup
  }{2015}]{song2015tvsum}
Song, Y.; Vallmitjana, J.; Stent, A.; and Jaimes, A.
\newblock 2015.
\newblock Tvsum: Summarizing web videos using titles.
\newblock In {\em CVPR},  5179--5187.

\bibitem[\protect\citeauthoryear{Song \bgroup et al\mbox.\egroup
  }{2016}]{song2016category}
Song, X.; Chen, K.; Lei, J.; Sun, L.; Wang, Z.; Xie, L.; and Song, M.
\newblock 2016.
\newblock Category driven deep recurrent neural network for video
  summarization.
\newblock In {\em ICMEW},  1--6.
\newblock IEEE.

\bibitem[\protect\citeauthoryear{Szegedy \bgroup et al\mbox.\egroup
  }{2015}]{szegedy2015going}
Szegedy, C.; Liu, W.; Jia, Y.; Sermanet, P.; Reed, S.; Anguelov, D.; Erhan, D.;
  Vanhoucke, V.; and Rabinovich, A.
\newblock 2015.
\newblock Going deeper with convolutions.
\newblock In {\em CVPR},  1--9.

\bibitem[\protect\citeauthoryear{Williams}{1992}]{williams1992simple}
Williams, R.~J.
\newblock 1992.
\newblock Simple statistical gradient-following algorithms for connectionist
  reinforcement learning.
\newblock {\em Machine learning} 8(3-4):229--256.

\bibitem[\protect\citeauthoryear{Xu \bgroup et al\mbox.\egroup
  }{2015}]{xu2015show}
Xu, K.; Ba, J.; Kiros, R.; Cho, K.; Courville, A.; Salakhudinov, R.; Zemel, R.;
  and Bengio, Y.
\newblock 2015.
\newblock Show, attend and tell: Neural image caption generation with visual
  attention.
\newblock In {\em ICML},  2048--2057.

\bibitem[\protect\citeauthoryear{Zhang \bgroup et al\mbox.\egroup
  }{2016a}]{zhang2016summary}
Zhang, K.; Chao, W.-L.; Sha, F.; and Grauman, K.
\newblock 2016a.
\newblock Summary transfer: Exemplar-based subset selection for video
  summarization.
\newblock In {\em CVPR},  1059--1067.

\bibitem[\protect\citeauthoryear{Zhang \bgroup et al\mbox.\egroup
  }{2016b}]{zhang2016video}
Zhang, K.; Chao, W.-L.; Sha, F.; and Grauman, K.
\newblock 2016b.
\newblock Video summarization with long short-term memory.
\newblock In {\em ECCV},  766--782.
\newblock Springer.

\bibitem[\protect\citeauthoryear{Zhao and Xing}{2014}]{zhao2014quasi}
Zhao, B., and Xing, E.~P.
\newblock 2014.
\newblock Quasi real-time summarization for consumer videos.
\newblock In {\em CVPR},  2513--2520.

\end{thebibliography}
\bibliographystyle{aaai}

\end{document}